\documentclass[a4paper, 10pt, conference]{ieeeconf}
\IEEEoverridecommandlockouts
\usepackage{eso-pic}
\usepackage[utf8]{inputenc}
\usepackage[T1]{fontenc}
\usepackage{url}
\usepackage{cite}
\usepackage{amsmath,amssymb,amsfonts}
\usepackage{graphicx}
\usepackage{textcomp}
\usepackage{xcolor}

\title{\LARGE \bf A Fully Automatic Pipeline for 3D Dendrite Instance Segmentation \\ in SBF-SEM*}

\author{Zewen Zhuo$^{1}$, Ilya Belevich$^{2}$, Eija Jokitalo$^{2}$,
Alejandra Sierra$^{1}$, and Jussi Tohka$^{1}$%
\thanks{*This work was supported by the Finnish Ministry of Education and Culture's Pilot for Doctoral Programmes (Pilot project Mathematics of Sensing, Imaging and
Modelling) and by the Research Council of Finland under the grant 358944 (Flagship of Advanced Mathematics for Sensing, Imaging and Modeling).}
\thanks{$^{1}$Z. Zhuo, A. Sierra, and J. Tohka are with the A.I. Virtanen Institute
for Molecular Sciences, University of Eastern Finland, Kuopio, Finland
(e-mail: zewen.zhuo@uef.fi, alejandra.sierralopez@uef.fi, jussi.tohka@uef.fi).}%
\thanks{$^{2}$I. Belevich and E. Jokitalo are with the Electron Microscopy Unit,
Institute of Biotechnology, University of Helsinki, Helsinki, Finland
(e-mail: ilya.belevich@helsinki.fi, eija.jokitalo@helsinki.fi).}%
\thanks{Corresponding authors: Z. Zhuo and J. Tohka.}%
}

\begin{document}
\AddToShipoutPictureFG*{%
  \AtPageUpperLeft{%
    \put(\LenToUnit{0.5\paperwidth},\LenToUnit{-0.5cm}){%
      \makebox[0pt][c]{\parbox[t]{0.85\paperwidth}{\centering\scriptsize
        \copyright~2026 IEEE. Personal use of this material is permitted. Permission
        from IEEE must be obtained for all other uses, in any current or future media,
        including reprinting/republishing this material for advertising or promotional
        purposes, creating new collective works, for resale or redistribution to servers
        or lists, or reuse of any copyrighted component of this work in other works.\\
        Accepted to be published in: Proceedings of the 2026 IEEE-EMBS Conference on
        Biomedical Engineering and Sciences (IECBES).}}}}}
\maketitle
\thispagestyle{empty}
\pagestyle{empty}

\begin{abstract}
Accurate three-dimensional (3D) reconstruction of individual dendrites in
serial block-face scanning electron microscopy (SBF-SEM) is essential for
quantifying structural plasticity in the brain, yet manual annotation at scale
is infeasible. We present a fully automatic pipeline for 3D dendrite instance
segmentation that unifies YOLOv6-guided Segment Anything Model (SAM) prompting
on downsampled slices, iterative two-dimensional mask refinement, random forest
3D instance linking, and instance-aware high-resolution refinement using nnU-Net
at native resolution into a single system requiring no manual prompting at
inference. Applied to hippocampal CA1 SBF-SEM datasets from a control rat and a
pilocarpine-induced epileptic rat, our pipeline reconstructs coherent,
well-separated dendrites with high semantic accuracy (Dice 0.93 and 0.91) and
strong instance-level performance on control tissue, while analysis of the more
challenging epileptic tissue identifies instance recognition in dense regions as
the principal remaining limitation. The high-resolution refinement stage recovers
thin dendritic protrusions, providing a basis for downstream spine-level
analysis. Code is available at \url{https://github.com/ZE-WEN/dendrite-3d-instance-seg}.

\vspace{1ex}
\textit{Clinical Relevance}--- Automated dendrite reconstruction enables scalable quantification of the dendritic and spine alterations seen in epilepsy models, which may help relate ultrastructural changes to disease mechanisms and treatment effects.
\end{abstract}


\section{Introduction}

Neuroscience increasingly relies on reconstructing three-dimensional (3D)
cellular and tissue architecture from stacks of two-dimensional (2D) sections,
enabling analyses that go beyond flat, single-section representations of
biological organization. Volumetric microscopy provides the necessary 3D
context, yet resolving individual structures as consistent 3D instances across
sections remains difficult wherever those structures are densely packed, thin,
or elongated. Dendrites and their spines illustrate this challenge concretely.
Dendrites integrate synaptic inputs, while their small protrusions, dendritic
spines, host most excitatory synapses and provide compartmentalized sites for
synaptic transmission and plasticity \cite{holtmaat2009,yuste1995}. Accurate
reconstruction of dendrites from volumetric electron microscopy (EM) is
thus valuable for studying structural remodeling in development,
plasticity, and neurological disease. In epilepsy, for instance, hippocampal CA1
dendrites show spine loss, varicose beading, and layer-selective shifts in spine
distribution \cite{postnikova2024,swann2000}. These changes yield denser and
more irregular dendritic profiles, posing challenges for generalizable methods.

Serial block-face scanning electron microscopy (SBF-SEM) provides
nanometer-scale in-plane resolution together with volumetric ultrastructural
information, making it suitable for analyzing dendritic shafts, spine heads, and
thin spine necks \textit{in situ} \cite{denk2004,peddie2022}. However, SBF-SEM
stacks pack neuronal and non-neuronal structures into large, anisotropic volumes
with weak or ambiguous boundaries. A reconstruction method must therefore
resolve each dendrite as a distinct 3D instance and track it consistently across
slices, which makes manual reconstruction difficult to scale.

Addressing this challenge has driven a shift toward learned segmentation, with
U-Net and nnU-Net as dominant approaches for semantic segmentation
\cite{ronneberger2015,isensee2021}. Recovering 3D instances from these
predictions remains a long-standing problem in EM, addressed by partitioning
boundary or affinity maps into neurites via agglomeration or multicut methods
\cite{beier2017,funke2019}, or through geometric post-processing
\cite{abdollahzadeh2021}. Flood-filling networks instead grow instances
iteratively \cite{januszewski2018}. These approaches produce accurate
reconstructions but rely on dense volumetric labels and task-specific training. Promptable foundation models provide a complementary direction. The Segment
Anything Model (SAM) generates masks from prompts such as points and bounding
boxes \cite{kirillov2023}, and its use in microscopy has been studied
\cite{archit2025}, including our own earlier work prompting SAM for dendrites
from single 2D slices without automatic detection or 3D linking
\cite{zhuo2025}. Detector-prompted variants remove manual interaction, as in
CellSAM for light-microscopy cells \cite{marks2025}. A parallel line of work
targets full 3D reconstruction by extending the memory mechanism of SAM-2
\cite{ravi2025}, propagating a prompted mask across slices through a learned
volumetric state. SAM4EM pairs a memory mechanism with a prompt-free adapter for
structures such as mitochondria, glia, and synapses \cite{shah2025}, and
Spatial-SAM replaces the memory mechanism of SAM-2 with a
signed-distance-function memory \cite{huang2026}. Both are optimized end to end,
so the propagation module and the segmentation backbone are co-adapted on the
target volume. Despite these efforts, existing 3D EM instance methods tie instance recovery to
either dense volumetric annotation with task-specific training or end-to-end
co-adaptation on the target volume. For dendrites, whose thin, branching
morphology makes dense 3D ground truth especially scarce, neither is practical.
How to reconstruct dendrite instances from modular, pretrained components,
without dense 3D annotation or joint end-to-end training, therefore remains open.

We propose a fully automatic 2D-to-3D pipeline for dendrite instance
reconstruction in SBF-SEM. In place of learned memory propagation, it resolves
inter-slice identity with an explicit, interpretable classifier and composes
modular, pretrained stages, avoiding joint end-to-end optimization. Our
contributions are fourfold:
\begin{itemize}
    \item A detector-prompted segmentation stage in which a YOLOv6 detector
    supplies bounding-box prompts to a promptable segmenter, giving fully
    automatic per-slice masks.
    \item An iterative 2D refinement loop coupled with a random forest
    (RF) inter-slice linking model that assembles slice-wise masks into coherent
    3D instances from engineered features.
    \item An instance-aware high-resolution refinement branch (nnU-Net at native
    resolution) that is identity-preserving, recovering thin
    protrusions attenuated by downsampling.
    \item An evaluation on control and epileptic hippocampal CA1 data that
    localizes the dominant failure mode to instance recognition in dense tissue rather than to delineation.
\end{itemize}

\section{Materials and Methods}

\subsection{Datasets and Annotations}

Two rat datasets from the CA1 region of the hippocampus were used. Dataset A was
acquired from a healthy rat and served as the control, and Dataset B was
acquired from a rat after pilocarpine-induced status epilepticus. Both were
imaged at an in-plane resolution of $15 \times 15$ nm with a 40 nm sectioning
interval, yielding anisotropic stacks of $3100 \times 3100$ pixels per slice.
After removing unusable border regions, the retained volumes spanned approximately
$46.5 \times 46.5 \times 35.7\ \mu\mathrm{m}^3$ (Dataset A, 8.6 GB
uncompressed) and $46.5 \times 46.5 \times 22.8\ \mu\mathrm{m}^3$ (Dataset B,
5.5 GB uncompressed). Animal procedures were approved by the Committee of the
Provincial Government of Southern Finland and followed European Community
Council Directives 86/609/EEC.

The main branch produces per-slice dendrite masks with a SAM variant fine-tuned
for slice-wise dendrite delineation \cite{zhuo2025}. Z.Z. created 2D sparse
object-mask annotations of Dataset A (100 annotated slices, 2168 2D dendrite
instances), and A.S., an expert in neuroanatomy, validated them. These annotations were
used to train the automatic YOLOv6 detector, taking each bounding-box label as
the enclosing rectangle of the corresponding mask. The RF 3D linking classifier
and the nnU-Net tissue classifier were trained on 30 slices of sparse 3D
instance annotation from Dataset A (27 3D instances). Specifically, we used the
inter-slice instance links derived from these labels for the RF classifier, and
for the nnU-Net we binarized the same 30 slices into background and dendrite
classes for 2D training. For evaluation, Z.Z. prepared sparse
3D instance-level annotations on both Dataset A and Dataset B (slices 452--461 of Dataset A and
285--294 of Dataset B). The Dataset A evaluation slices did not overlap with any slice used to train YOLOv6, the fine-tuned SAM, the RF linker, or nnU-Net, and no Dataset B slice was used at any training stage.

\subsection{Preprocessing}

Before segmentation, the stacks were aligned to reduce inter-slice displacement
using the StackReg plugin in ImageJ \cite{thevenaz1998} on downsampled $1024 \times 1024$ stacks. The slice-wise transformations were then scaled and applied
to the original-resolution slices, giving consistent stacks at both resolutions.
Two aligned representations were used. For YOLO-guided SAM segmentation the
stack was downsampled to $1024 \times 1024$ pixels, while the
original-resolution stack was retained for high-resolution refinement.

\subsection{YOLO-Guided SAM Segmentation}

\begin{figure*}[t]
\centerline{\includegraphics[width=0.86\textwidth]{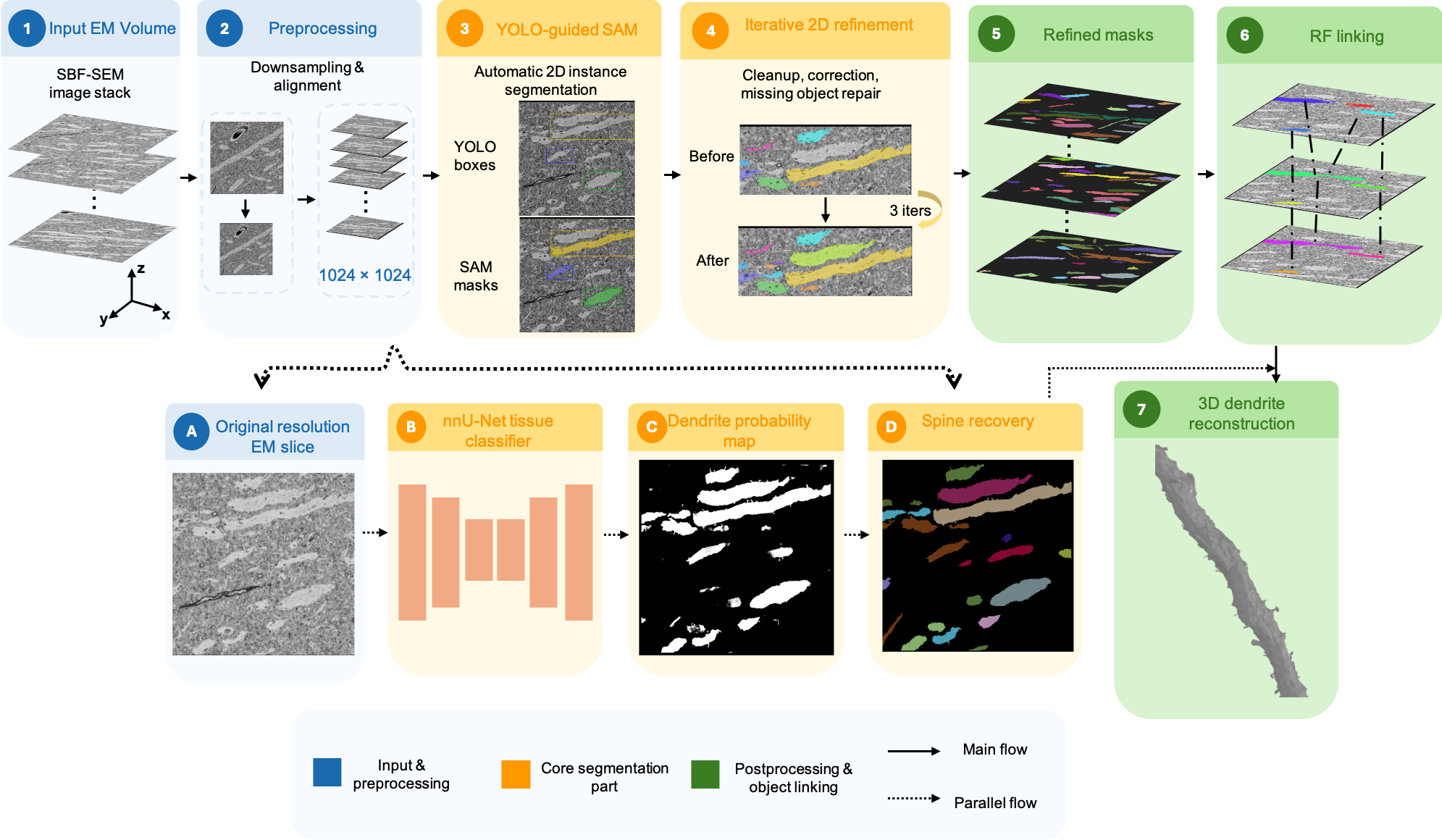}}
\caption{Overview of the proposed pipeline. The main branch (1--7) produces and
links 2D instance masks from downsampled slices. The high-resolution branch
(A--D) applies nnU-Net at original resolution for protrusion recovery, and the
two branches merge at step 7.}
\label{fig1}
\end{figure*}

The pipeline operates in two coordinated branches, as shown in Fig.~\ref{fig1}.
We select YOLOv6-L \cite{li2022} to balance model capacity and inference
efficiency and train it for 400 epochs using stochastic gradient descent with a
cosine learning-rate schedule, combining GIoU box-regression loss, distribution
focal loss, and a classification loss. The detector is applied to the
downsampled and aligned EM slices to detect candidate dendrites in each slice.
Each predicted bounding box is used as a separate prompt to SAM, producing one
binary mask per box. The resulting per-box masks are then assembled into a
single 2D instance map for the slice.

\subsection{Iterative 2D Mask Refinement}

Since SAM processes each bounding-box prompt independently, initial masks contain artifacts such as spurious components, holes, missed detections, and split labels. We reduce these through an iterative procedure alternating between mask post-processing and supplementary SAM re-prompting.

In each iteration, masks are post-processed slice by slice, where each non-background label is decomposed into 4-connected components, components with area smaller than 500 pixels are removed, and all enclosed holes are filled. A harmonization step then evaluates nearby component pairs from different labels using four criteria, namely spatial proximity, normalized contact strength, overlap after dilation, and background gap thickness, merging pairs that satisfy the criteria. Additionally, foreground regions present in slices $z+1$ and $z+2$ but absent in slice $z$ are converted into supplementary bounding-box prompts for the next SAM iteration. This cycle is iterated three times.

\subsection{3D Instance Linking}

Given the refined 2D instance mask stack, an RF classifier \cite{breiman2001}
assigns a linking probability to each candidate inter-slice component pair,
converting independent slice-wise masks into coherent volumetric dendrite
instances.

Training data are derived from the sparse 3D instance annotations, in which each
dendrite carries a consistent label across slices. Within each slice, every
annotated instance is split into its 2D connected components, and candidate
links are formed between components on adjacent slices whose pixel overlap
exceeds a minimum threshold. A link is labeled positive when both components
belong to the same instance.

For each candidate link between components $c_i^{z}$ and $c_j^{z+1}$ on adjacent
slices $z$ and $z+1$, we extract an eight-dimensional feature vector
\begin{equation}
f(c_i^{z},c_j^{z+1})=[\,o_{ij},\,\mathrm{IoU}_{ij},\,A_i,\,A_j,\,r_{ij},\,\Delta y,\,\Delta x,\,d_{ij}\,],
\end{equation}
where $o_{ij}=|c_i^{z} \cap c_j^{z+1}|$ is the number of overlapping pixels and
$A_i$, $A_j$ are the component areas. The remaining features are the
intersection over union (IoU), $\mathrm{IoU}_{ij}=o_{ij}/(A_i+A_j-o_{ij})$, the
area ratio $r_{ij}=\min(A_i,A_j)/\max(A_i,A_j)$, the centroid displacement
$(\Delta y, \Delta x)$, and the Euclidean centroid distance
$d_{ij}=\|\mu_i-\mu_j\|_2$, with $\mu_i$ and $\mu_j$ the component centroids. At
inference, the trained classifier assigns a linking probability to each
candidate edge between adjacent slices, and edges with probability above a
threshold $p_{\mathrm{th}}$ are retained. Retained edges are resolved into 3D
instances with a union-find pass over the component graph, so a dendrite that
branches within the volume is kept as a single connected instance instead of being split at the branch point.

\subsection{High-Resolution Refinement}

Downsampling reduces the effective in-plane resolution from approximately $15
\times 15$ nm$^2$ to approximately $45 \times 45$ nm$^2$ per pixel, weakening thin spine necks
and fine protrusions. To recover these structures, nnU-Net \cite{isensee2021} is
applied at the original resolution after RF-based 3D linking. The downsampled
instance segmentation is upscaled to the original resolution by nearest-neighbor
interpolation, which preserves discrete instance identities. Independently,
nnU-Net produces a high-resolution dendrite probability volume for the whole
stack. The recovery step combines the two, using the probability volume as
tissue evidence and the upscaled instances to assign recovered voxels to the
correct dendrite.

We train a binary 2D nnU-Net on the 30 annotated original-resolution slices.
nnU-Net self-configures the training setup from the dataset properties,
selecting Z-score normalization, a patch size of $1280 \times 1024$, and a batch
size of 2.

Recovery is performed on all linked instances. Each instance is cropped to its
3D bounding box, and within each slice the probability volume is thresholded at
0.98 and split into 4-connected components. Any component overlapping the
instance on that slice is taken as evidence for the same dendrite and added to
its mask. Because recovery is additive, unioning components without removing
voxels, the instance identity is retained while thin, high-confidence
protrusions are reattached at full resolution. After recovery, a set of post-processing steps is applied to consolidate the final 3D instances. Full implementation details are released as open-source code.

\section{Experiments and Results}

\subsection{Evaluation Protocol}

We evaluated on Dataset A (control) and Dataset B (epileptic), each on 10
sparsely annotated slices. Dense 3D instance ground truth (GT) is infeasible to
produce manually at scale, so each reference was annotated sparsely. To reduce
annotation time, the pipeline output was used as an initialization. Each
instance was then manually refined pixel by pixel on every annotated slice, and
dendrites missed by the pipeline were annotated from scratch. This limits the
dependence of the reference on the method, although some bias toward the
pipeline output, mainly at ambiguous boundaries, cannot be fully excluded and
was accepted as a trade-off against annotation time. The low instance-level
scores on Dataset B further indicate that the reference did not simply inherit
the pipeline's instance decisions.

Evaluation followed a sparse GT strategy consistent with the general
recommendations of \cite{maierhein2024}, in which the union of annotated
reference voxels defined the evaluated region, and predicted instances lying
entirely outside it were ignored rather than counted as false positives, since
they might be valid but unannotated structures. This strategy underlies all
metrics defined below.

All metrics were computed on 3D label volumes following established
recommendations for biomedical image segmentation evaluation
\cite{maierhein2024,muller2022}. For a GT instance $g$ and prediction $p$,
$\mathrm{IoU}(g,p)=|g \cap p|/|g \cup p|$ and $\mathrm{Dice}(g,p)=2|g \cap
p|/(|g|+|p|)$. At a threshold $t$, a bipartite graph was used to keep only pairs
with $\mathrm{IoU} \geq t$, and a one-to-one matching by the Hungarian algorithm
\cite{kuhn1955} maximized total IoU, giving $TP(t)$, $FP(t)$, and $FN(t)$.
Following \cite{caicedo2019}, we summarized instance quality by
$S_{\mathrm{inst}}$, the mean instance Jaccard score over the thresholds
$T=\{0.10,0.15,\ldots,0.90\}$,
\begin{equation}
S_{\mathrm{inst}}=|T|^{-1}\sum_{t \in T}\frac{TP(t)}{TP(t)+FP(t)+FN(t)}.
\end{equation}

We additionally report the single-threshold scores $S_{\mathrm{inst}}@0.5$ and
$S_{\mathrm{inst}}@0.75$, where $S_{\mathrm{inst}}@t$ denotes the instance
Jaccard evaluated at the single threshold $t$ rather than averaged over $T$. As
a panoptic summary \cite{kirillov2019} at $t=0.5$, segmentation quality (SQ) is
the mean IoU over matched pairs, detection quality (DQ) is $2TP/(2TP+FP+FN)$,
and panoptic quality is $\mathrm{PQ}=\mathrm{SQ} \times \mathrm{DQ}$. For
semantic quality, all instances were collapsed into binary foreground masks
$G=\bigcup_g g$ and $P=\bigcup_p p$, treating every labeled voxel as foreground.
Because the annotation was sparse and foreground-only, a predicted voxel outside
the annotated region could not be adjudicated as an error and was left
unpenalized. Precision is therefore not identifiable, and the semantic scores
reduce to the recovery of annotated dendrite voxels. Dice and IoU were computed
as above with $g=G$ and $p=P$, using $TP=|G \cap P|$ and $FN=|G \setminus P|$ as
voxel counts, so the reported IoU equals the fraction of annotated dendrite
voxels recovered.

\begin{table}[t]
\caption{Quantitative results on Dataset A (control) and Dataset B (epileptic).}
\label{tab1}
\centering
\renewcommand{\arraystretch}{1.15}
\begin{tabular}{llcc}
\hline
Category & Metric & Dataset A & Dataset B \\
\hline
Instance-level & $S_{\mathrm{inst}}$ & 0.63 & 0.38 \\
 & $S_{\mathrm{inst}}@0.5$ & 0.75 & 0.34 \\
 & $S_{\mathrm{inst}}@0.75$ & 0.44 & 0.11 \\
\hline
Panoptic & PQ & 0.70 & 0.37 \\
 & SQ & 0.82 & 0.72 \\
 & DQ & 0.86 & 0.51 \\
\hline
Sparse semantic & Dice & 0.93 & 0.91 \\
 & IoU & 0.87 & 0.83 \\
\hline
\end{tabular}
\end{table}
\begin{figure*}[t]
\centerline{\includegraphics[width=\textwidth]{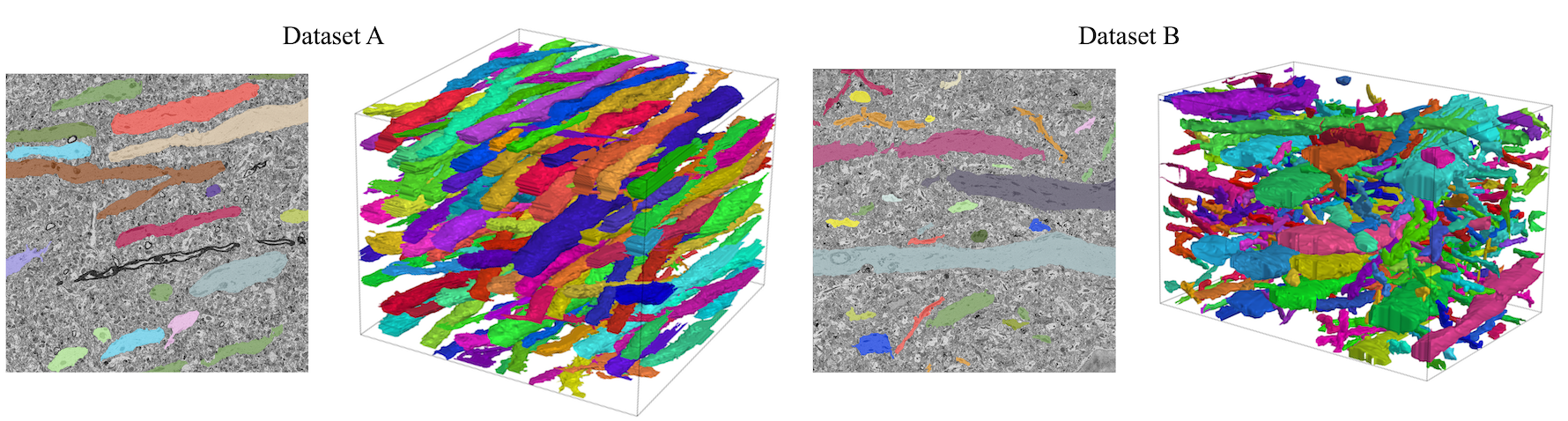}}
\caption{Qualitative results on Dataset A (control) and Dataset B (epileptic).
The left panel of each pair overlays predicted dendrite instances, each in a
distinct color, on a single original-resolution EM slice. The right panel shows
the corresponding 3D instance reconstruction over the whole volume.}
\label{fig2}
\end{figure*}

\subsection{Quantitative Results}

Table~\ref{tab1} reports the instance-level, panoptic, and sparse semantic
metrics. Semantic recovery of annotated dendrite voxels was high and stable
across tissue types (Dice 0.93 and 0.91, IoU 0.87 and 0.83), so almost all
annotated dendrite voxels were recovered in both conditions. Segmentation
quality was high on control tissue and moderate on epileptic tissue (SQ 0.82 and
0.72), so correctly matched instances were well delineated on control tissue and
acceptably delineated on the harder diseased tissue.

The instance-level and detection metrics, however, fell markedly from Dataset A
to Dataset B. Detection quality dropped from 0.86 to 0.51 and, since
$\mathrm{PQ}=\mathrm{SQ} \times \mathrm{DQ}$, drove the panoptic decline (0.70
to 0.37), with the instance score following ($S_{\mathrm{inst}}$ 0.63 to 0.38).
Because SQ declined far less than DQ, the dominant errors lay in recognizing and
separating dendrites rather than in delineating them. Two factors were
consistent with this. The detector, trained only on control tissue, likely had
lower recall on denser epileptic profiles. Therefore, missed detections yielded poor SAM
prompts and suppressed recognition without affecting delineation. Instances
detected only on their largest cross-sections, with thin leading and trailing
slices missed, fell below the 3D IoU threshold, lowering DQ and
$S_{\mathrm{inst}}$, while their recovered voxels still contributed to the
semantic scores.

The instance scores are conservative by construction: $S_{\mathrm{inst}}$
averages instance Jaccard over IoU thresholds up to 0.90. Thus,  a few-voxel
disagreement on thin branching structures can drop an otherwise correct instance
below the threshold ($S_{\mathrm{inst}}@0.75$ of 0.44 and 0.11). Large gaps
between pixel-level Dice and threshold-based scores are therefore expected
\cite{caicedo2019}.

\begin{table}[t]
\caption{Ablation of the inter-slice linking model.}
\label{tab2}
\centering
\renewcommand{\arraystretch}{1.15}
\begin{tabular}{lcccc}
\hline
 & \multicolumn{2}{c}{Dataset A} & \multicolumn{2}{c}{Dataset B} \\
\cline{2-3} \cline{4-5}
Metric & RF & IoU only & RF & IoU only \\
\hline
$S_{\mathrm{inst}}$ & \textbf{0.63} & 0.33 & \textbf{0.38} & 0.25 \\
$S_{\mathrm{inst}}@0.5$ & \textbf{0.75} & 0.29 & \textbf{0.34} & 0.30 \\
$S_{\mathrm{inst}}@0.75$ & \textbf{0.44} & 0.29 & \textbf{0.11} & 0.08 \\
\hline
\end{tabular}
\end{table}

\subsection{Ablation of the Inter-Slice Linking Model}

The RF linker introduces a learned component into an otherwise geometric
assembly step and we ask whether the learned scoring is necessary. We compared the RF linker
against a purely geometric baseline that removes the classifier and establishes
inter-slice correspondence from raw overlap alone, operating on the same refined
2D instance masks. For each pair of consecutive slices the baseline computes the
IoU between every pair of components and accepts a link when the two components
are mutual best matches and their IoU exceeds 0.6, and accepted links are resolved
into 3D instances by the same union-find pass.

Table~\ref{tab2} reports the comparison. The RF linker outperformed the geometric
baseline on both datasets, with the largest margin on Dataset A (healthy tissue,
$S_{\mathrm{inst}}$ 0.63 against 0.33). The gap narrowed on Dataset B (epileptic tissue,
$S_{\mathrm{inst}}$ 0.38 against 0.25), where both linkers were constrained by the upstream
detection failures described above rather than by the scoring of individual
edges. On control tissue the baseline scores were identical at
$S_{\mathrm{inst}}@0.5$ and $S_{\mathrm{inst}}@0.75$ (0.29 in both cases), so
the instances it did assemble were delineated accurately and its loss came from
tracks it failed to assemble at all, not from imprecise ones. As a result, learned scoring contributed most where recognition was otherwise reliable, by
sustaining tracks through the slice-to-slice variation that a fixed overlap
criterion rejects.

Although RF linkage outperformed the geometric baseline, it comes with a practical limitation. Six of its eight features are expressed in absolute pixel units, so a classifier trained at one working resolution cannot be applied at another without retraining, whereas an overlap ratio is scale-free. Two properties of the baseline IoU method also bound the comparison. It matches components one to one, so a dendrite that branches within the volume is split at the branch point, and it uses a single overlap threshold that we did not tune. The result consequently establishes that the learned scorer helps at the operating point tested, not that no geometric rule could match it.

\subsection{Qualitative Results}

Fig.~\ref{fig2} shows representative qualitative results for both datasets. For
the control tissue (Dataset A), reconstructed dendrites form coherent, clearly
separated shafts. In the epileptic tissue (Dataset B), the arrangement is denser
and more irregular, with a larger proportion of small, fragmented profiles that
are harder to detect and track consistently across slices.

\section{Discussion and Conclusion}

We presented a fully automatic pipeline for 3D dendrite instance segmentation in
SBF-SEM, integrating automatic detection, promptable 2D segmentation, iterative
refinement, RF 3D linking, and instance-aware high-resolution refinement into a
single system that operates without manual prompting at inference. On control
tissue it reconstructs coherent, well-separated dendrites with high semantic
accuracy (Dice 0.93) and strong instance-level performance. On the epileptic
tissue, sparse semantic recovery remains high and segmentation quality moderate while
detection and instance-level scores fall, which locates the dominant errors in
recognizing and separating dendrites rather than in delineating them. An
ablation of the linking stage shows that the learned edge scorer is a
substantive component instead of an incidental one, since replacing it with
fixed-threshold overlap matching roughly halves the instance score on control
tissue. Volume EM resolves fine neurites at nanometer scale but is restricted to
small volumes by acquisition time and data size \cite{peddie2022}. Reducing the
manual effort per EM volume, as pursued here, extends nanometer-scale
reconstruction to more tissue in the meantime.

These results come from one animal per condition, and the RF linker and nnU-Net were trained on only 30 annotated slices of Dataset A, so generalization to other animals and acquisition protocols remains to be established. Transfer is nonetheless plausible, because no stage was trained on epileptic tissue, yet segmentation quality declined far less than detection on Dataset B, and the RF linker relies on geometric rather than appearance features. The modularity of the pipeline points to several concrete next steps. We attribute the performance drop to detection and not delineation, which leads to the adaptation of the detector, either by fine-tuning YOLOv6 on epileptic tissue or by training a separate detector per condition. Validation on more animals and on densely annotated subvolumes, which would also allow precision to be measured, is needed to confirm these results. Another direction extends the method toward its intended downstream use. Because the
high-resolution stage reattaches thin protrusions at native resolution, these
protrusions provide the substrate on which individual dendritic spines can be
identified and separated, enabling the spine-level morphological analysis this
reconstruction is meant to support.

\section*{Acknowledgment}

 The authors wish to acknowledge UEF Bioinformatics Center and CSC -- IT Center for Science, Finland, for computational resources. We thank Mohammad
Khateri and Melina Estela Dalmau for their meaningful insights.

\bibliographystyle{IEEEtran}
\bibliography{references}

@article{holtmaat2009,
  title={Experience-dependent structural synaptic plasticity in the mammalian brain},
  author={Holtmaat, Anthony and Svoboda, Karel},
  journal={Nature Reviews Neuroscience},
  volume={10},
  number={9},
  pages={647--658},
  year={2009},
  publisher={Nature Publishing Group UK London}
}

@article{yuste1995,
  title={Dendritic spines as basic functional units of neuronal integration},
  author={Yuste, Rafael and Denk, Winfried},
  journal={Nature},
  volume={375},
  number={6533},
  pages={682--684},
  year={1995},
  publisher={Nature Publishing Group UK London}
}

@article{postnikova2024,
  title={Morphological and functional alterations in the {CA1} pyramidal neurons of the rat hippocampus in the chronic phase of the lithium--pilocarpine model of epilepsy},
  author={Postnikova, Tatyana Y and Diespirov, Georgy P and Malkin, Sergey L and Chernyshev, Alexander S and Vylekzhanina, Elizaveta N and Zaitsev, Aleksey V},
  journal={International Journal of Molecular Sciences},
  volume={25},
  number={14},
  pages={7568},
  year={2024},
  publisher={MDPI}
}

@article{swann2000,
  title={Spine loss and other dendritic abnormalities in epilepsy},
  author={Swann, John W and Al-Noori, Salwa and Jiang, Minghui and Lee, Chong L},
  journal={Hippocampus},
  volume={10},
  number={5},
  pages={617--625},
  year={2000},
  publisher={Wiley Online Library}
}

@article{denk2004,
  title={Serial block-face scanning electron microscopy to reconstruct three-dimensional tissue nanostructure},
  author={Denk, Winfried and Horstmann, Heinz},
  journal={PLoS Biology},
  volume={2},
  number={11},
  pages={e329},
  year={2004},
  publisher={Public Library of Science San Francisco, USA}
}

@article{peddie2022,
  title={Volume electron microscopy},
  author={Peddie, Christopher J and Genoud, Christel and Kreshuk, Anna and Meechan, Kimberly and Micheva, Kristina D and Narayan, Kedar and Pape, Constantin and Parton, Robert G and Schieber, Nicole L and Schwab, Yannick and others},
  journal={Nature Reviews Methods Primers},
  volume={2},
  number={1},
  pages={51},
  year={2022},
  publisher={Nature Publishing Group UK London}
}

@inproceedings{ronneberger2015,
  title={{U-Net}: Convolutional networks for biomedical image segmentation},
  author={Ronneberger, Olaf and Fischer, Philipp and Brox, Thomas},
  booktitle={International Conference on Medical Image Computing and Computer-Assisted Intervention (MICCAI)},
  pages={234--241},
  year={2015},
  organization={Springer}
}

@article{isensee2021,
  title={{nnU-Net}: a self-configuring method for deep learning-based biomedical image segmentation},
  author={Isensee, Fabian and Jaeger, Paul F and Kohl, Simon AA and Petersen, Jens and Maier-Hein, Klaus H},
  journal={Nature Methods},
  volume={18},
  number={2},
  pages={203--211},
  year={2021},
  publisher={Nature Publishing Group US New York}
}

@article{beier2017,
  title={Multicut brings automated neurite segmentation closer to human performance},
  author={Beier, Thorsten and Pape, Constantin and Rahaman, Nasim and Prange, Timo and Berg, Stuart and Bock, Davi D and Cardona, Albert and Knott, Graham W and Plaza, Stephen M and Scheffer, Louis K and others},
  journal={Nature Methods},
  volume={14},
  number={2},
  pages={101--102},
  year={2017},
  publisher={Nature Publishing Group}
}

@article{funke2019,
  title={Large scale image segmentation with structured loss based deep learning for connectome reconstruction},
  author={Funke, Jan and Tschopp, Fabian and Grisaitis, William and Sheridan, Arlo and Singh, Chandan and Saalfeld, Stephan and Turaga, Srinivas C},
  journal={IEEE Transactions on Pattern Analysis and Machine Intelligence},
  year={2019},
  volume={41},
  number={7},
  pages={1669-1680},
  publisher={IEEE}
}

@article{abdollahzadeh2021,
  title={{DeepACSON} automated segmentation of white matter in {3D} electron microscopy},
  author={Abdollahzadeh, Ali and Belevich, Ilya and Jokitalo, Eija and Sierra, Alejandra and Tohka, Jussi},
  journal={Communications Biology},
  volume={4},
  number={1},
  pages={179},
  year={2021},
  publisher={Nature Publishing Group UK London}
}

@article{januszewski2018,
  title={High-precision automated reconstruction of neurons with flood-filling networks},
  author={Januszewski, Micha{\l} and Kornfeld, J{\"o}rgen and Li, Peter H and Pope, Art and Blakely, Tim and Lindsey, Larry and Maitin-Shepard, Jeremy and Tyka, Mike and Denk, Winfried and Jain, Viren},
  journal={Nature Methods},
  volume={15},
  number={8},
  pages={605--610},
  year={2018},
  publisher={Nature Publishing Group US New York}
}

@inproceedings{kirillov2023,
  title={Segment anything},
  author={Kirillov, Alexander and Mintun, Eric and Ravi, Nikhila and Mao, Hanzi and Rolland, Chloe and Gustafson, Laura and Xiao, Tete and Whitehead, Spencer and Berg, Alexander C and Lo, Wan-Yen and others},
  booktitle={2023 IEEE/CVF International Conference on Computer Vision (ICCV)},
  pages={3992--4003},
  year={2023},
  organization={IEEE}
}

@article{archit2025,
  title={Segment anything for microscopy},
  author={Archit, Anwai and Freckmann, Luca and Nair, Sushmita and Khalid, Nabeel and Hilt, Paul and Rajashekar, Vikas and Freitag, Marei and Teuber, Carolin and Spitzner, Melanie and Tapia Contreras, Constanza and others},
  journal={Nature Methods},
  volume={22},
  number={3},
  pages={579--591},
  year={2025},
  publisher={Nature Publishing Group US New York}
}

@inproceedings{zhuo2025,
  title={Segment anything for dendrites from electron microscopy},
  author={Zhuo, Zewen and Belevich, Ilya and Leinonen, Ville and Jokitalo, Eija and Malm, Tarja and Sierra, Alejandra and Tohka, Jussi},
  booktitle={2025 IEEE 6th International Conference on Image Processing, Applications and Systems (IPAS)},
  pages={1--6},
  year={2025},
  organization={IEEE}
}

@article{marks2025,
  title={{CellSAM}: a foundation model for cell segmentation},
  author={Marks, Markus and Israel, Uriah and Dilip, Rohit and Li, Qilin and Yu, Changhua and Laubscher, Emily and Iqbal, Ahamed and Pradhan, Elora and Ates, Ada and Abt, Martin and others},
  journal={Nature Methods},
  volume={22},
  number={12},
  pages={2585--2593},
  year={2025},
  publisher={Nature Publishing Group US New York}
}

@inproceedings{ravi2025,
  title={{SAM} 2: Segment anything in images and videos},
  author={Ravi, Nikhila and Gabeur, Valentin and Hu, Yuan-Ting and Hu, Ronghang and Ryali, Chaitanya and Ma, Tengyu and Khedr, Haitham and R{\"a}dle, Roman and Rolland, Chloe and Gustafson, Laura and others},
  booktitle={International Conference on Learning Representations (ICLR)},
  year={2025}
}

@inproceedings{shah2025,
  title={{SAM4EM}: Efficient memory-based two stage prompt-free segment anything model adapter for complex {3D} neuroscience electron microscopy stacks},
  author={Shah, Uzair and Agus, Marco and Boges, Daniya and Chiappini, Vanessa and Alzubaidi, Mahmood and Schneider, Jens and Hadwiger, Markus and Magistretti, Pierre J and Househ, Mowafa and Cal{\'\i}, Corrado},
  booktitle={2025 IEEE/CVF Conference on Computer Vision and Pattern Recognition Workshops (CVPRW)},
  pages={4678--4687},
  year={2025},
  organization={IEEE}
}

@inproceedings{huang2026,
  title={{Spatial-SAM}: Spatially Consistent {3D} Electron Microscopy Segmentation with {SDF} Memory and Semi-Supervised Learning},
  author={Huang, Yikai and Han, Renmin and Wang, Yuxuan and Cai, Youcheng and Liu, Ligang},
  booktitle={Proceedings of the IEEE/CVF Conference on Computer Vision and Pattern Recognition (CVPR)},
  pages={22763--22772},
  year={2026}
}

@article{thevenaz1998,
  title={A pyramid approach to subpixel registration based on intensity},
  author={Thevenaz, Philippe and Ruttimann, Urs E and Unser, Michael},
  journal={IEEE Transactions on Image Processing},
  volume={7},
  number={1},
  pages={27--41},
  year={1998},
  publisher={IEEE}
}

@article{li2022,
  title={{YOLOv6}: A single-stage object detection framework for industrial applications},
  author={Li, Chuyi and Li, Lulu and Jiang, Hongliang and Weng, Kaiheng and Geng, Yifei and Li, Liang and Ke, Zaidan and Li, Qingyuan and Cheng, Meng and Nie, Weiqiang and others},
  journal={arXiv preprint arXiv:2209.02976},
  year={2022}
}

@article{breiman2001,
  title={Random forests},
  author={Breiman, Leo},
  journal={Machine Learning},
  volume={45},
  number={1},
  pages={5--32},
  year={2001},
  publisher={Springer}
}

@article{maierhein2024,
  title={Metrics reloaded: recommendations for image analysis validation},
  author={Maier-Hein, Lena and Reinke, Annika and Godau, Patrick and Tizabi, Minu D and Buettner, Florian and Christodoulou, Evangelia and Glocker, Ben and Isensee, Fabian and Kleesiek, Jens and Kozubek, Michal and others},
  journal={Nature Methods},
  volume={21},
  number={2},
  pages={195--212},
  year={2024},
  publisher={Nature Publishing Group US New York}
}

@article{muller2022,
  title={Towards a guideline for evaluation metrics in medical image segmentation},
  author={M{\"u}ller, Dominik and Soto-Rey, I{\~n}aki and Kramer, Frank},
  journal={BMC Research Notes},
  volume={15},
  number={1},
  pages={210},
  year={2022},
  publisher={Springer}
}

@article{kuhn1955,
  title={The {Hungarian} method for the assignment problem},
  author={Kuhn, Harold W},
  journal={Naval Research Logistics Quarterly},
  volume={2},
  number={1-2},
  pages={83--97},
  year={1955},
  publisher={Wiley Online Library}
}

@article{caicedo2019,
  title={Nucleus segmentation across imaging experiments: the 2018 {Data Science Bowl}},
  author={Caicedo, Juan C and Goodman, Allen and Karhohs, Kyle W and Cimini, Beth A and Ackerman, Jeanelle and Haghighi, Marzieh and Heng, CherKeng and Becker, Tim and Doan, Minh and McQuin, Claire and others},
  journal={Nature Methods},
  volume={16},
  number={12},
  pages={1247--1253},
  year={2019},
  publisher={Nature Publishing Group US New York}
}

@inproceedings{kirillov2019,
  title={Panoptic segmentation},
  author={Kirillov, Alexander and He, Kaiming and Girshick, Ross and Rother, Carsten and Doll{\'a}r, Piotr},
  booktitle={2019 IEEE/CVF Conference on Computer Vision and Pattern Recognition (CVPR)},
  pages={9396--9405},
  year={2019},
  organization={IEEE}
}

\end{document}